\documentclass[letterpaper, 10 pt, conference]{ieeeconf-modified}  %

\IEEEoverridecommandlockouts                              %

\usepackage[numbers]{natbib}
\usepackage{graphicx}
\graphicspath{ {./figures} }
\usepackage{caption}
\usepackage{siunitx}
\usepackage{algorithm}
\usepackage{algorithmic}
\usepackage{amsmath} %
\usepackage{amssymb}  %
\usepackage{textcomp}
\usepackage{hyperref}
\usepackage{booktabs}
\usepackage{tablefootnote}

\usepackage[capitalize]{cleveref}
\crefname{equation}{}{}

\title{\LARGE \bf
Dextrous Tactile In-Hand Manipulation Using a \\Modular Reinforcement Learning Architecture %
}
\author{Johannes Pitz, Lennart Röstel, Leon Sievers and Berthold Bäuml%
}

\begin{document}

\twocolumn[{%
        \renewcommand\twocolumn[1][]{#1}%
        \maketitle
        \begin{center}
            \centering
            \captionsetup{type=figure}
            \vskip -0.3cm
            \includegraphics[width=0.85\textwidth]{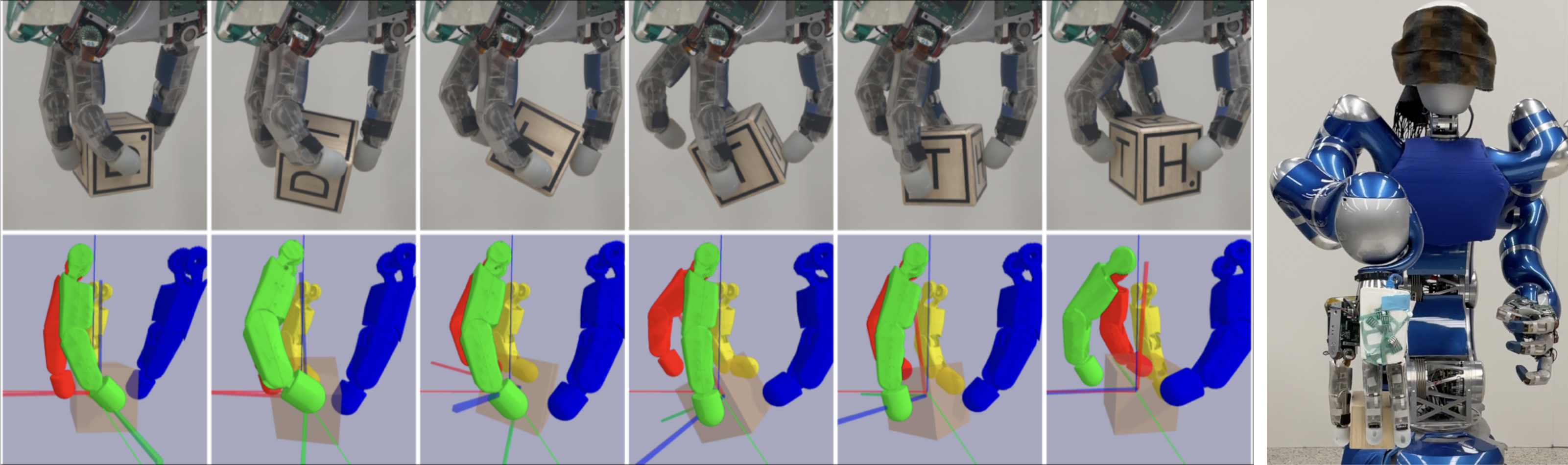}
            \captionof{figure}{The torque-controlled DLR-Hand II~\cite{Butterfass2001} performing in-hand manipulation. The cube is rotated to a goal orientation which can be reached by three $\pi/2$ rotations (see \cref{fig:24_goals} for all 24 goal orientations; here start in 3, end in 20). The task is performed purely tactile without external sensors (no cameras -- Agile Justin~\cite{Bauml2014} is blindfolded). The lower row shows the estimated cube state from the trained deep differentiable particle filter.}
            \label{fig:sequence}
        \end{center}%
    }]
{\let\thefootnote\relax\footnote[0]{The authors are with the DLR Institute of Robotics and Mechatronics, Technical University of Munich, and Deggendorf Institute of Technology. \newline
Contact:{\tt\scriptsize{ \{johannes.pitz|berthold.baeuml\}@dlr.de}}\newline}}

\thispagestyle{empty}
\pagestyle{empty}

\vskip -0.3cm
\begin{abstract}
    Dextrous in-hand manipulation with a multi-fingered robotic hand is a challenging task, esp. when performed with the hand oriented upside down, demanding permanent force-closure, and when no external sensors are used.
    For the task of reorienting an object to a given goal orientation (vs. infinitely spinning it around an axis), the lack of external sensors is an additional fundamental challenge as the state of the object has to be estimated all the time, e.g., to detect when the goal is reached.
    In this paper, we show that the task of reorienting a cube to any of the 24 possible goal orientations in a $\pi/2$-raster using the torque-controlled DLR-Hand II is possible.
    The task is learned in simulation using a modular deep reinforcement learning architecture:
    the actual policy has only a small observation time window of 0.5\,s but gets the cube state as an explicit input which is estimated via a deep differentiable particle filter trained on data generated by running the policy.
    In simulation, we reach a success rate of 92\% while applying significant domain randomization.
    Via zero-shot Sim2Real-transfer on the real robotic system, all 24 goal orientations can be reached with a high success rate.
    {(Web:~{\footnotesize\url{dlr-alr.github.io/dlr-tactile-manipulation}})}
\end{abstract}

\section{Introduction}

Many important application domains, like industrial manufacturing or housework, still has still to be done by humans. 
One important reason for this is the need for dextrous fine manipulation.
In particular, the task of in-hand manipulation, i.e., reorienting an object inside a hand, is often needed but challenging due to the intricate multi-contact dynamics, the coordination of the many degrees of freedom (DOF) of a multi-fingered hand, and the estimation of the state of the manipulated object.
In this paper, we extend our previous work~\cite{Sievers2022} on learning purely tactile in-hand manipulation with a torque-controlled hand from being able to rotate a cube around a single axis, to reorienting the cube to any of the 24 possible orientations in a $\pi/2$-raster (see \cref{fig:sequence}).
We solve the task in a realistic setting by holding the hand upside down, hence, demanding permanent force closure, and without external sensors (like cameras) but using only the fingers' integrated position and torque sensors.
The task is learned from scratch using a modular deep reinforcement learning architecture where learning the manipulation strategy is separated from learning an estimator for the cube state. \cref{fig:setting_hand_cube} and \cref{fig:architecture} give an overview of the robotic and task setup as well as the control architecture.

\subsection{Related Work}

In a seminal work, \citet{OpenAI2018} presented for the first time dextrous in-hand manipulation of a cube on a hand facing upward.
In order to execute the goal-oriented policy, the necessary cube state was estimated from images of multiple cameras.
Recently, \citet{nvidia2022dextreme} showed real-world results on the same task with a simpler camera setup.
In contrast to this, in the setting we study here, permanent force closure has to be kept because of the hand pointing downward.
Additionally, we use only the hand's integrated position and torque sensors without visual feedback, which is more realistic but also creates a significantly harder state estimation problem.

In our previous work~\cite{Sievers2022}, we used the same setting as presented here, but the task was to rotate the cube continuously around the vertical axis.
And \citet{Qi2022} even showed to do so with diverse objects.
For the task we present here, the object not only has to be rotated around an arbitrary axis but also a state estimator for tracking the cube state is required as a specific goal orientation has to be reached.
Purely in simulation, \citet{Khandate2022} presented tactile in-hand manipulation where simple geometric shapes could be continuously rotated around the x-, y-, and z-axes while keeping force closure. Also only in simulation and with state estimation from visual input, \citet{Chen2021} have shown goal-based reorientation of diverse objects with the hand upside down in some experiments. 
For a more extensive discussion of the large body of work on in-hand manipulation in simulation, we refer to our previous paper~\cite{Sievers2022}.

\subsection{Contributions}

\begin{itemize}
    \item We show that purely tactile (using only torque and position sensors) in-hand manipulation for reorienting a cube to any of the 24 goal orientations is possible. 
    \item We devise a modular learning architecture, separately training the policy for the actual finger control using the cube state as part of the observations, and a differentiable particle filter-based estimator~\cite{Rostel2022a} using only torque and position sensor readings.
    \item We show how we iteratively refine policy and estimator, modify the reward, and utilize domain randomization and asymmetric observations to successfully combine both modules.
   \item In simulation, a success rate of 92\% is reached despite significant domain randomization. With zero-shot Sim2Real transfer, all goal orientations are reached on the real DLR-Hand II.
\end{itemize}

\begin{figure}
    \centering
    \includegraphics[width=0.8\linewidth]{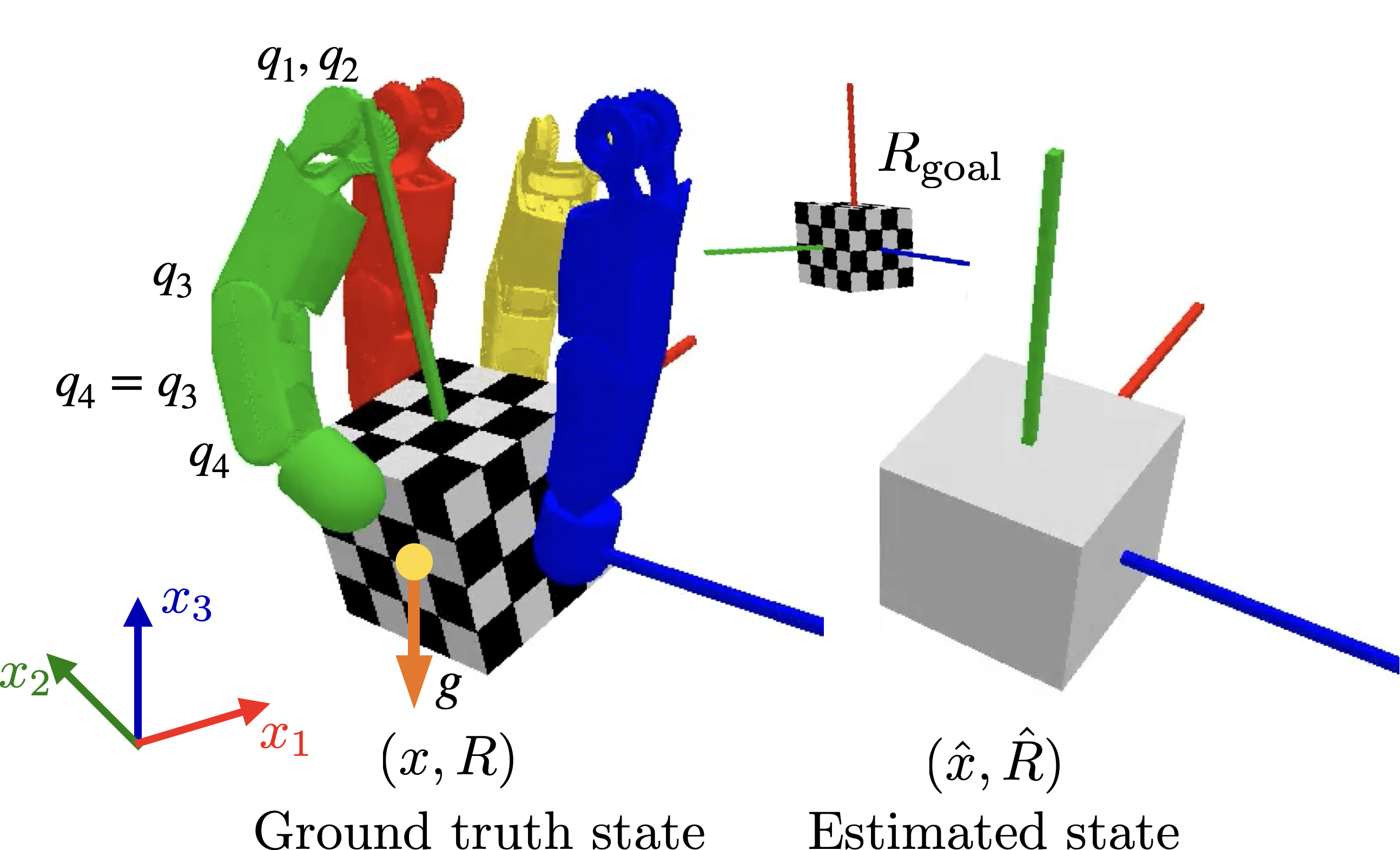}
    \caption{
        Robot and task setting. Each finger of the four-fingered hand has three active $q_1, q_2, q_3$ and one passive joint $q_4$.
        The cube should be rotated from a start orientation to a goal orientation $R_\text{goal}$. %
        We define the (positive) angle between the current orientation $R$ and the goal orientation $R_\text{goal}$ as $\theta = d(R_\text{goal}, R)$.
        During training steps S1-S3 (cf. \cref{tab:training}) the ground truth cube state is provided to the policy, whereas during testing a deep particle filter is used to estimate the state.
    }
    \label{fig:setting_hand_cube}
    \vskip -0.5cm
\end{figure}

\section{Modular Learning Architecture}

\subsection{Motivation}
Although the trend in machine learning is to learn everything end-to-end in one giant neural network model, sometimes it is still useful to modularize the architecture, e.g., to gain insights into the system during execution and especially during learning.
To solve the in-hand manipulation task described above the controller needs to keep track of the object state during execution.
Otherwise, it would not be possible to distinguish different multiple $\pi/2$ orientations of the cube.
Now, it could be possible to train a recurrent network controller to solve this task in simulation, but if the policy does not work on the real robot it seems next to impossible to analyze what is going wrong.

\begin{figure}
    \vspace{3mm}
    \centering
    \includegraphics[width=0.9\linewidth]{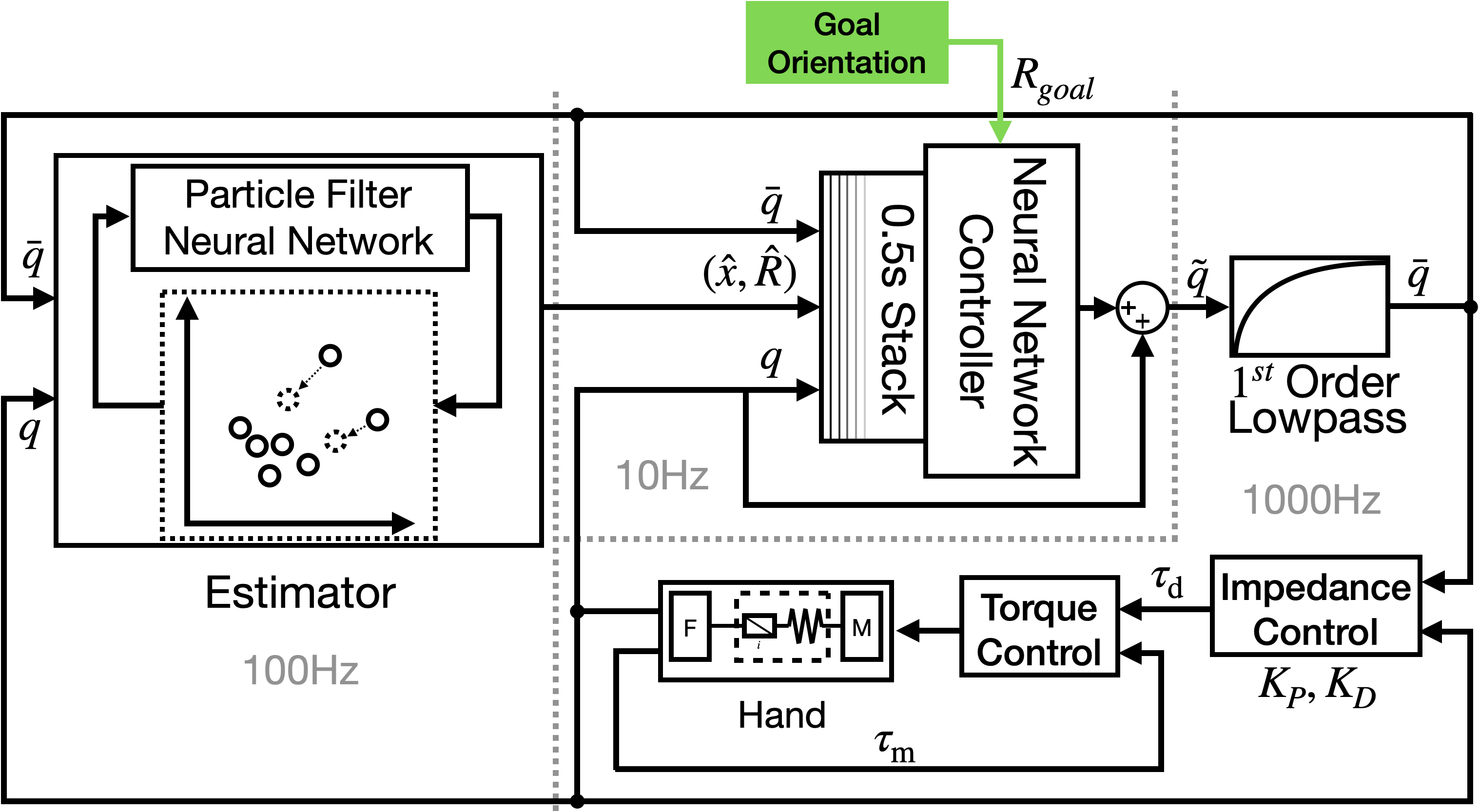}
    \caption{
        Overall system architecture.
        The DLR-Hand II~\cite{Butterfass2001} provides joint-level high-fidelity impedance control by using the integrated torque sensors, acting as a configurable spring-damper system.
        By sending desired joint angles as a unifying interface, free space motion as well as detailed control of the forces in contact can be realized at a moderate update rate (10\,Hz) of the policy network.
        Via a lowpass filter, the dynamics of the system are explicitly reduced to ease the Sim2Real transfer.
        The input of the policy network is calculated from the desired goal orientation, the measured joint angles as well the estimated cube pose and stacked over the last 0.5~seconds.
        The cube state (position and orientation) is estimated by a learned differentiable particle filter using the measured and the commanded finger angles, this way it is getting information about contacts.
    }
    \label{fig:architecture}
    \vskip -0.5cm
\end{figure}

\subsection{Modules}

The overall architecture is depicted in~\cref{fig:architecture} and comprises the following components (details are described later in the respective sections).
\subsubsection{State Estimator}
When considering the necessary information for performing goal-oriented reorientation, the state of the cube is an obvious choice for the abstraction layer and has been used as such in in-hand manipulation~\cite{OpenAI2018,Chen2021,OpenAI2019}.
But unlike these works, we want to estimate the object's state in a purely tactile manner.
Thereto, we employ a variant of differentiable particle filters~\cite{Jonschkowski2018,Karkus2018}, a learning based-method which we have proven to work in the context of in-hand manipulation in \citet{Rostel2022a}.

\subsubsection{Network Controller}
Relying on a separate module to estimate the cube state allows using a simple multi-layer perceptron (2 layers with 512 units) for the controller network (i.e., without memory like an LSTM network).
\citet{Sievers2022} showed that such a network can learn complex in-hand manipulation tasks even with a short time horizon of 0.5\,s.
The advantages are that the network is easier to train and investigate in predefined configurations.

\section{Learning the Controller}

The controller network is trained as the policy in the reinforcement learning framework.
We use the off-policy algorithm Soft Actor-Critic (SAC)~\cite{Haarnoja2018}. %
Our custom implementation of the algorithm and the simulation using the physics engine PyBullet~\cite{Coumans2016} already proved to be suitable for real-world robotics~\cite{Sievers2022}.

The overall control architecture is summarized in \cref{fig:architecture}.
Note that in our prior work, we limited the range of the fingers by working with smaller joint ranges.
Now, for the more intricate goal rotation task, we work with the full range of motion.
And we found that due to the increased range, we had to modify the controller such that the policy $\pi$ outputs angles relative to the current joint angles (instead of directly outputting the pre-filter desired joint angles $\tilde{q}$).
\begin{equation}
    \tilde{q}_{t+1} = \mathrm{clip}(q_t + \pi(o_t) \frac{\tau_\text{max}}{K_\text{p}}, q_\text{min}, q_\text{max})
\end{equation}
The full observation $o_t$ passed to the policy is shown in \cref{tab:observation} and explained in~\cref{subsection:observaton}.
$K_\text{p}$ is a constant of the underlying impedance controller and $\tau_\text{max}$ the maximally allowed torque.

Moreover, we now use 120 workers running the simulation to collect data instead of 12.
Increasing the number of workers helps to cope with a wider range of domain randomization.
To make use of the off-policy nature of SAC we also increase the replay buffer size by a factor of 10 to 1.5 million steps.
All details of the training including the learning and simulation parameters can be found on accompanying website.

\subsection{Learning Environment}

Although the task we want to solve is to reorient a cube into any of 24 possible $\pi/2$ goal orientations (cf.~\cref{fig:24_goals}), during training we continue to sample new goal orientations once the current one is reached.
Learning with this task should improve Sim2Real performance by preventing short-sighted behavior and increasing robustness towards initial conditions.

\begin{table}
    \vspace{3mm}
    \caption{Observation}
    \label{tab:observation}
    \centering
    \scriptsize
    \begin{tabular}{ccc}
        \toprule
        Name                 & Q-function                                & Policy                                   \\
        \midrule
        Joint angles         & \multicolumn{2}{c}{$q_t$}                                                            \\
        Desired angles       & \multicolumn{2}{c}{$\bar{q}_{t-1}$}                                                  \\
        Control error        & \multicolumn{2}{c}{$\bar{q}_{t-1} - q_t$}                                            \\ %
        Goal orientation     & \multicolumn{2}{c}{$R_{\text{goal}}$}                                              \\
        Cube state           & $(x_t, R_{\text{sym}, t})$              & $(\hat{x}_t, \hat{R}_{\text{sym}, t})$ \\
        Delta rotation       & $R_\text{goal}^{-1}R_t$                 & $R_\text{goal}^{-1}\hat{R}_t$          \\ %
        Cube linear velocity & $v_t = \dot{x}_t$                         &                                          \\
        \bottomrule
    \end{tabular}
    \caption*{
        Note: The complete input to the neural networks consists of a time stack of length $S=5$ of the respective quantities shown above.
        To exploit the symmetries of the cube, its orientation~$R$ is reduced with the octahedral group to $R_\text{sym}$.
        Rotations are passed in as quaternions.
    }
    \vskip -0.5cm
\end{table}

\subsubsection{Initial state}
The environment is reset by setting a random orientation for the cube and placing it on top of another fixed cube.
Then the hand closes, the same way it can be done on the real system and the fixed cube disappears. After a few more simulation steps we use this state as the first observation for the learning algorithm.

\subsubsection{Termination}
An episode ends in a failure if the cube drops down (reaches -5\,cm in $x_3$ direction) or moves more than 10\,cm away from the origin.
Successfully reaching the goal is defined as holding the cube within $|x| < 2.5\text{\,cm}$ and delta orientation angle $\theta < 0.4\text{\,rad}$ (cf.~\cref{fig:setting_hand_cube}) for 400\,ms (4 update steps). %
Only these conditions generate a termination signal for the learning algorithm.
Additionally, we reset the environment without sending a termination signal if a goal is not reached after 10\,s or if a total of 120\,s has passed, to ensure that the replay buffer always contains data from different instances of the domain randomization.

\subsection{Reward}

During training (cf.~\cref{subsection:training}) we use two different reward functions.
Initially, we use the "goal" reward function $r_\text{g}$ introduced by~\citet{Chen2021}.
\begin{equation*}
    r_\text{g}=    \frac{\lambda_\theta}{\theta + \epsilon_\theta} - \mathrm{clip}(\lambda_\text{pos} \|x\| ^ 4, 0, \lambda_\text{clip}) +
    \begin{cases}
        \lambda_\text{drop}, & \text{if drop} \\
        \lambda_\text{succ}, & \text{if succ.} \\
        0,                     & \text{else.}
    \end{cases}
\end{equation*}
The first term rewards small delta orientation angle $\theta$, with some $\epsilon_\theta$ to avoid singularities.
The second term penalizes deviations in the cube position.
Reaching the goal or dropping the cube is explicitly included.
Different terms are weighted by $\lambda$ factors and we omit the time index $t$ when it is clear from the context.

Later we switch to a "simpler" reward funtion $r_\text{s}$ based on the relative change of the position $\Delta x_t = \|x_t\| - \|x_{t-1}\|$ and delta orientation $\Delta \theta_t =  \theta_t - \theta_{t-1}$.
\begin{equation}
    r_\text{s}=\mathrm{clip}(-\lambda_\theta' \Delta \theta, -\infty, \lambda_\text{clip}') - \lambda_\text{pos}' \Delta x.
\end{equation}
Note that we do not explicitly reward success or drop events.
Successfully reaching goals is desirable for the policy due to the dense reward and dropping the cube is penalized by the position component of the reward.

The main advantages of the simpler reward function are better interpretability of the final episode reward (because the physical quantities enter linearly), and clipping the maximal reward per step forces the policy to prioritize generalizing across the randomized domains versus optimizing for high reward in specific environment configurations.
In the future, we want to avoid using the goal reward completely.

\subsection{Observations}
\label{subsection:observaton}

Thanks to the modular learning architecture we can train the policy independently from the estimator.
However, that increases the importance of the modeled sensor noise.
Often asymmetric observations are used to exploit privileged information in the form of additional observations to the Q-function (or Value-function)~\cite{Sievers2022, OpenAI2018, OpenAI2019}.
In this work, we separate the policy and Q-function observations completely, such that we can pass ground truth states to the Q-function and (potentially extremely) noisy signals to the policy.
Additionally, only the Q-function receives the linear cube velocities.
See \cref{tab:observation} for a complete overview.

\subsection{Domain Randomization}
\label{subsection:DR}

For a successful Sim2Real transfer we apply domain randomization during simulation. See~\cref{tab:DR} for the most important parameters and the website for a complete list.
\subsubsection{Sensor noise}
Sensor noise is sampled at each step, all other randomizations take place before a new episode starts.
We apply gaussian noise to the joint angles and the cube state.
But note, we add sensor noise only to the policy inputs (cf. \cref{subsection:observaton}) and we do not add additional noise if the policy receives cube states predicted by the estimator~(i.e. S4, S5 in \cref{tab:training} and the benchmark).
\subsubsection{Controller}
We use sticky actions \cite{Machado2018} to imitate communication delays on the real system.
We sample $K_p$, $K_d$, parasitic stiffness, and joint angle offsets $q_\text{off}$ as identified in~\citet{Sievers2022}.
\subsubsection{Cube}
We sample the cube's mass, size, and initial pose.
We apply small random forces and torques to the cube.
\subsubsection{Friction}
It is crucial to randomize friction parameters (lateral $\eta_\text{lat}$ and spinning $\eta_\text{spin}$ for finger links and cube).
See \cref{subsection:spinning_friction} for an explanation and the empirical impact of the spinning friction on the policy's performance.

\begin{table}
    \vspace{3mm}
    \caption{Domain randomization}
    \label{tab:DR}
    \centering
    \scriptsize
    \begin{tabular}{c c c}
        \toprule
        Parameter              & Distribution\tablefootnote{
            $\mathrm{Log}\mathcal{U}$ refers to the loguniform distribution.
        }                      & Notes\tablefootnote{
            Shows if we switch off or fix the randomization during some training stages, cf.~\cref{tab:training}.
            B refers to the benchmark (\cref{subsection:protocol}).
        }                                                                                           \\
        \midrule
        \multicolumn{3}{c}{Per Step}                                                                \\
        \midrule
        $q$ [rad]              & $\mathcal{N}(0, 0.02)$                            &                \\
        $x$ [m]                & $\mathcal{N}(0, 0.01)$                            & S4, S5, B: off \\
        $R$ [rad]              & $\mathcal{N}(0, 0.2)$                             & S4, S5, B: off \\
        \midrule
        \multicolumn{3}{c}{Per Episode}                                                             \\
        \midrule
        $q_\text{off}$ [rad] & $\mathcal{U}(-0.04, 0.04)$                        &                \\
        $\eta_\text{lat}$    & $\mathcal{U}(0.81, 0.99)$                         &                \\
        $\eta_\text{spin} \left[\frac{\mathrm{Nm}}{\mathrm{N}}\right]$\tablefootnote{
            Given our parametrization, PyBullet computes the effective spinning friction between the fingertips and the cube as $\eta_\text{spin}  \eta_\text{lat} / 0.9$.
        }                      & $\mathrm{Log}\mathcal{U}(\num{2e-4}, \num{2e-2})$ & B: fixed       \\
        \bottomrule
    \end{tabular}
    \vskip -0.5cm
\end{table}

\subsection{Trainnig Procedure and Curriculum}
\label{subsection:training}

The policy used in the experiment sections and the accompanying video was trained in a multi-step procedure,
see~\cref{tab:training} for an overview and the policies' performance on the benchmark (\cref{subsection:protocol}).
Initially (S1, S2) the policy receives the true cube states from the simulator, with added noise, and later (S5) we pass in the cube states that the estimator predicts at the time (cf.~\cref{subsection:estimator_in_the_loop}).

During the first step (S1) we trained on the goal reward $r_\text{g}$ and use a curriculum that slowly increases the gravity until it reaches $g$ (common for in-hand manipulation~\cite{Sievers2022,Chen2021}).

Every time we train a new policy we use all the weights (and optimizer states) to initialize a fresh replay buffer and continue learning in a slightly modified setup.
In S2 we change the reward function and increase the filter constant of the first-order lowpass (to slow down the dynamics, which is easier to simulate and safer to execute in the real world).
In S5 we switch the policy input to use the predicted cube states.

The total training time was around two weeks on machines with up to 80 cores.
Details are on the project website. %

\subsection{Estimator In-the-Loop}
\label{subsection:estimator_in_the_loop}

Given a policy that works well in simulation and a fully trained estimator (cf.~\cref{section:estimator}), we perform one last training to fine-tune the policy to work with estimated states.
We run the estimator along the simulation and pass the estimated cube states to the policy (without added noise), while the Q-function still receives the true states.
Additionally, we add a term to the reward function to incentivize the policy to perform actions that lead to predictable state transitions.
To that end, in the "estimator" reward $r_\text{e}$, we punish estimation errors in position and orientation.
\begin{equation}
    r_\text{e} = r_\text{s} + \mathrm{clip}(\lambda_\text{pos}''\|\hat{x} - x\|^2 +\lambda_\phi \phi^2, 0, \lambda_\text{clip}'')
\end{equation}
with $\phi = d(\hat{R}, R)$ (the angle between the estimated and true orientation).
We terminate episodes if $\|\hat{x} - x\| > \text{ 1.5\,cm}$ or $\phi > \text{0.8\,rad}$ to ensure that there won't be episodes in the replay buffer where the estimator is, for example, $\pi/2$ off, such that the manipulation works as expected but a lot of negative reward is accumulated.

\section{Learning the State Estimator}
\label{section:estimator}

The hand-object system is fully described by considering the state of the hand and the state of the cube.
Because joint positions and velocities are measured directly through proprioceptive sensors, we are only concerned with estimating the state of the cube
\begin{equation}
    s_t = (x_t, R_t, v_t, w_t)^T \in \mathbb{R}^3 \times \mathrm{SO}(3) \times \mathbb{R}^3 \times \mathbb{R}^3
\end{equation}
with cartesian position $x_t$, orientation $R_t$, lateral velocity $v_t$ and rotational velocity $w_t$.

Estimating the state of the system only from tactile feedback is especially challenging because no global information about the manipulated object is available.
We identify two main sources of uncertainty:
\begin{itemize}
    \item Uncertain dynamics, arising from the imprecisely modeled, contact-based interactions between the hand and the object (such as friction effects).
    \item Ambiguities of the object shape itself, that prevent uniquely determining the pose of the object solely from tactile information.
\end{itemize}
Bayesian filtering methods may be used to systematically account for these uncertainties, however, the uncertain, highly non-linear dynamics preclude the use of analytical or simulated dynamic models.
We employ D2P2F~\cite{Rostel2022a}, a differentiable particle filter~\cite{Jonschkowski2018,Karkus2018} variant that was shown to be suitable for purely tactile state estimation.

Particle filters keep a set of particles $\{(w^{(i)}, s^{(i)})\}_{i=1, \ldots, N}$  with weights and states. 
Differentiable particle filters learn expressive functions that are applied to the particles, implementing the prediction and update steps within the Bayesian filtering framework.
Specifically, D2P2F learns a generative proposal distribution $F_\varphi(\cdot | s_{t-1}^{(i)}, z_t, u_t)$ and an update model $G_\varphi(s_{t-1}^{(i)}, z_t, u_t)$, where $z_t = (q_t, \dot{q}_t)$ is an observation vector, consisting of the measured joint angles and joint velocities, and $u_t$ is the control input, consisting of the desired joint angles $\bar{q}_t$. The estimator is updated with a rate of $100$\,Hz (cf. \cref{fig:architecture}).
We refer to~\citet{Rostel2022a} for a more detailed description of D2P2F.

\subsection{Training}

We take the predicted state $\hat{s}_t$ as the composition of the weighted mean of particle positions, the medoid of particle rotations, and the weighted mean of lateral and rotational velocities.
We optimize for the weights of $F_\varphi$ and $G_\varphi$ of the D2P2F by minimizing the loss function 
\begin{equation}
    \label{eq:filter_loss}
    \mathcal{L}_\varphi = \frac{1}{T} \sum_{t=1}^{T} \sum_j c_j d_j(\hat{s}_{j,t}, s_{j,t})^2,
\end{equation}
for a trajectory of $T$ timesteps, where $d_j(\cdot)$ is the distance function associated with the $j$-th components of $s$; the euclidian distance for cartesian dimensions and the angle between two orientations respectively.
The weighting coefficients $c_j$ account for different scales of dimensions. We set $c_x=1.0$,  $c_R=100.0$ and  $c_v=c_w=0.1$.
The filter is trained on rollouts of the policy in simulation in a supervised manner.

The training is conducted in three stages:
First, the proposal model is trained for 1-step-ahead prediction ($T=1$) with a single particle until convergence.
Then, the filter is trained on sequences of $T=100$ timesteps by backpropagation through time.
For this, initial particles are sampled from a Normal distribution $\mathcal{N}(s_0, \sigma)$, where $\sigma$ corresponds to the standard deviation of the training data and $s_0$ is the true initial state plus a small bias term during training.
Finally, additional training data is generated in simulation where the policy receives predictions from the current model as input.
The newly collected \textit{in-loop} data is appended to the original dataset and the model is trained for 2 epochs. After that, the filter model used for data collection is updated.
This process is repeated iteratively until the fraction of \textit{in-loop} to offline data samples is 1/2, at which point we found that generating more \textit{in-loop} samples does not decrease prediction errors.

\section{Experiments}

\begin{table}
    \vspace{3mm}
    \caption{Training seqeunce}
    \label{tab:training}
    \centering
    \scriptsize
    \begin{tabular}{c p{53mm} c}
        \toprule
        Step & Training                                                                                               & rate $b$ \\
        \midrule
        S1   & Train policy $\pi_0'$ on reward $r_\text{g}$ with true cube state.                                   & $0.68$   \\
        S2   & Refine policy on reward $r_\text{s}$, resulting in $\pi_0.$                                          & $0.99$   \\
        S3   & Train an estimator $f_0$ on data generated from $\pi_0$.                                               & $0.74$   \\
        S4   & Iteratively refine an estimators $f_i$ on data from $\pi_0$ and cube state from $f_{i-1}$ in-the-loop. & $0.76$   \\
        S5   & Refine policy on reward $r_\text{e}$ to $\pi_1$ with $f_i$ in-the-loop.                                              & $0.92$   \\
        \bottomrule
    \end{tabular}
    \vskip -0.5cm
\end{table}

\subsection{Benchmark Protocol}

\label{subsection:protocol}

Since we have no external tracking system we cannot compute rewards on the real robot.
Instead, we check if the combination of estimator and controller can reach all 24 $\pi/2$ orientations of the cube.

The cube is passed in by a human the same way we initialize episodes in simulation.
Then, a goal is selected and the robot has to rotate the cube to match the goal orientation.
If the cube drops the episode is stopped and deemed unsuccessful.

When we benchmark a policy in simulation we perform 8 runs for each combination of the 24 goals and 3 spinning friction values ($\eta_\text{spin} =\num{2e-4}, \num{1e-3}, \num{1e-2}$).
We fix the spinning friction because it strongly influences the result.
Moreover, we fix the cube size to its nominal size of 8\,cm, again due to the high influence (whilst the real cube can be measured accurately).
Other domain randomization is still active.

\subsection{Evaluating the Policy}

\begin{figure*}
    \centering
    \vspace{2mm}
    \includegraphics[width=1.0\textwidth]{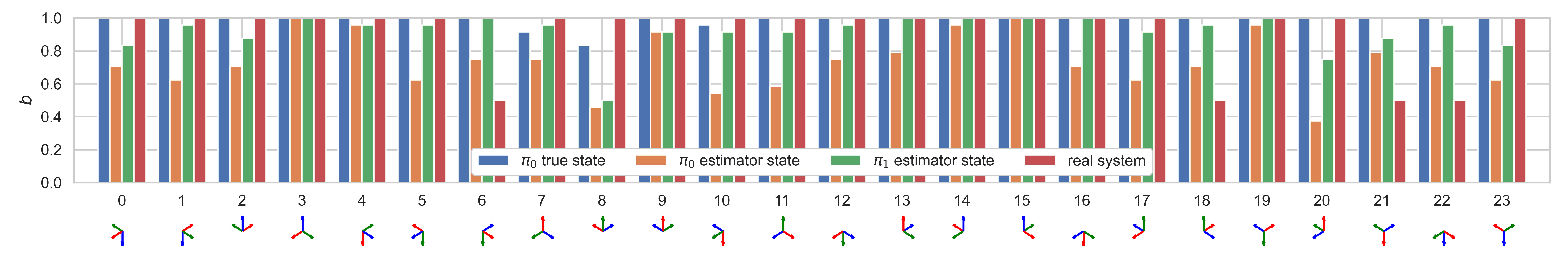}
    \vskip -0.3cm
    \caption{
        Average success rates $b$ for each of the goal orientations with visualizations of all 24 multiple $\pi/2$ orientations (the octahedral group). Note that goal 3 is the initial orientation of the cube (i.e. does not require a rotation).
        See~\cref{subsection:protocol} for details regarding the testing protocol.
    }
    \label{fig:24_goals}
    \vskip -0.5cm
\end{figure*}

In~\cref{tab:training} we detail the training steps and report intermediate success rates on the benchmark.
After the first step, the policy is already able to reach all goals but not very reliably.
After fine-tuning on the "simple" reward function, the policy reaches almost 100\% success rate, even if executed with noisy (simulator) states.
However, taking the estimator in-the-loop reduces the performance drastically.
Iteratively refining the estimator improves the performance slightly.
Finally, fine-tuning the policy on the latest estimator iteration shows a great improvement on the benchmark.
We also show the success rates for each goal individually in~\cref{fig:24_goals} along with the results on the real robot.
On the real system, we performed 4 runs for each goal, hence, the results are only qualitative.

\subsection{Evaluating the Estimator}
\begin{figure}
    \vspace{2mm}
    \centering
    \includegraphics[width=0.99\linewidth]{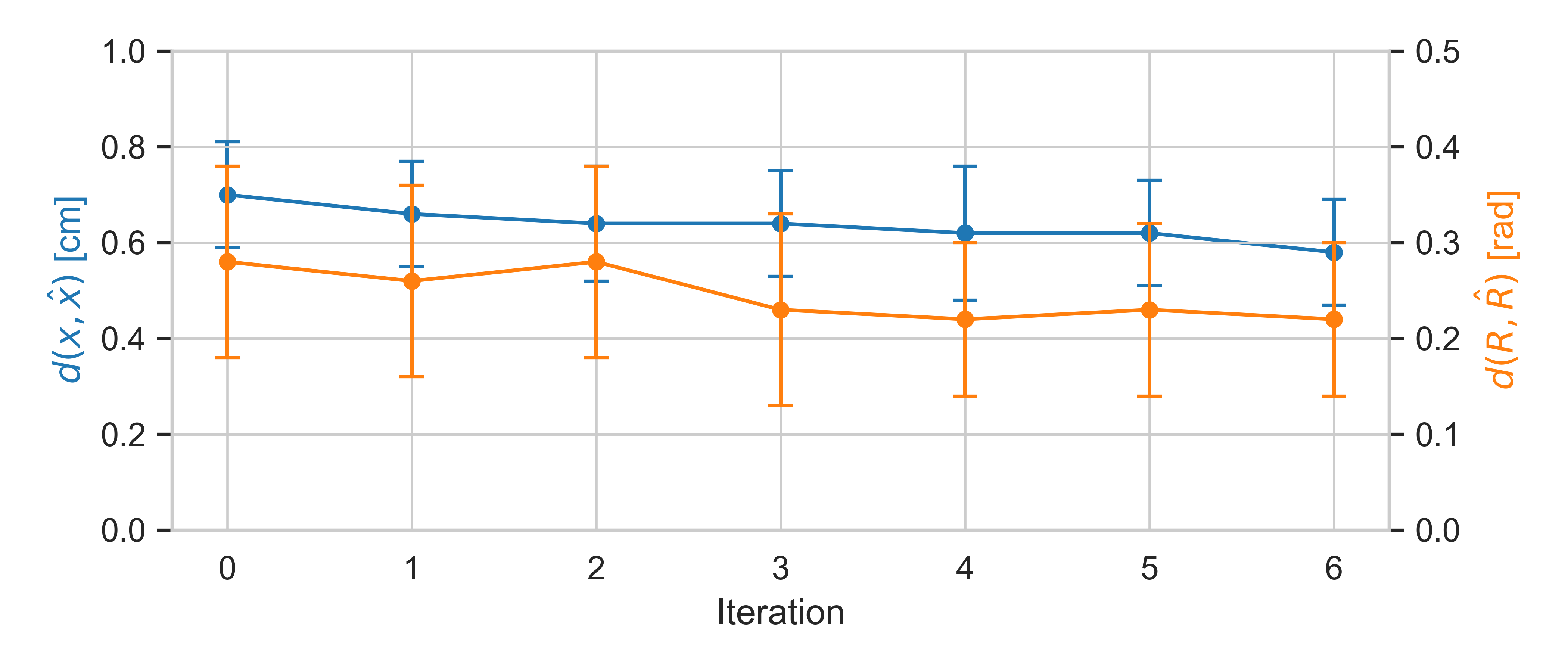}
    \vskip -0.3cm
    \caption{Position prediction error $d(x, \hat{x})$, and rotational prediction error $d(R, \hat{R})$ during iterative training of the estimator \textit{in-loop} with the policy.}
    \label{fig:filter_iterations_error}
    \vskip -0.3cm
\end{figure}

We evaluate the prediction accuracy of the state estimator during in-the-loop rollouts with different parameters of domain randomization. As shown in \cref{fig:filter_iterations_error}, prediction errors decrease over the iterative training procedure, although the estimator was trained until convergence on the data generated in the offline setting. This confirms the importance of training the estimator on data that is generated in conjunction with the policy.
However, note that the improvement in prediction accuracy only results in a minor performance improvement on the benchmark task (compare S3, S4 in \cref{tab:training}).

\begin{figure}
    \centering
    \includegraphics[width=0.99\linewidth]{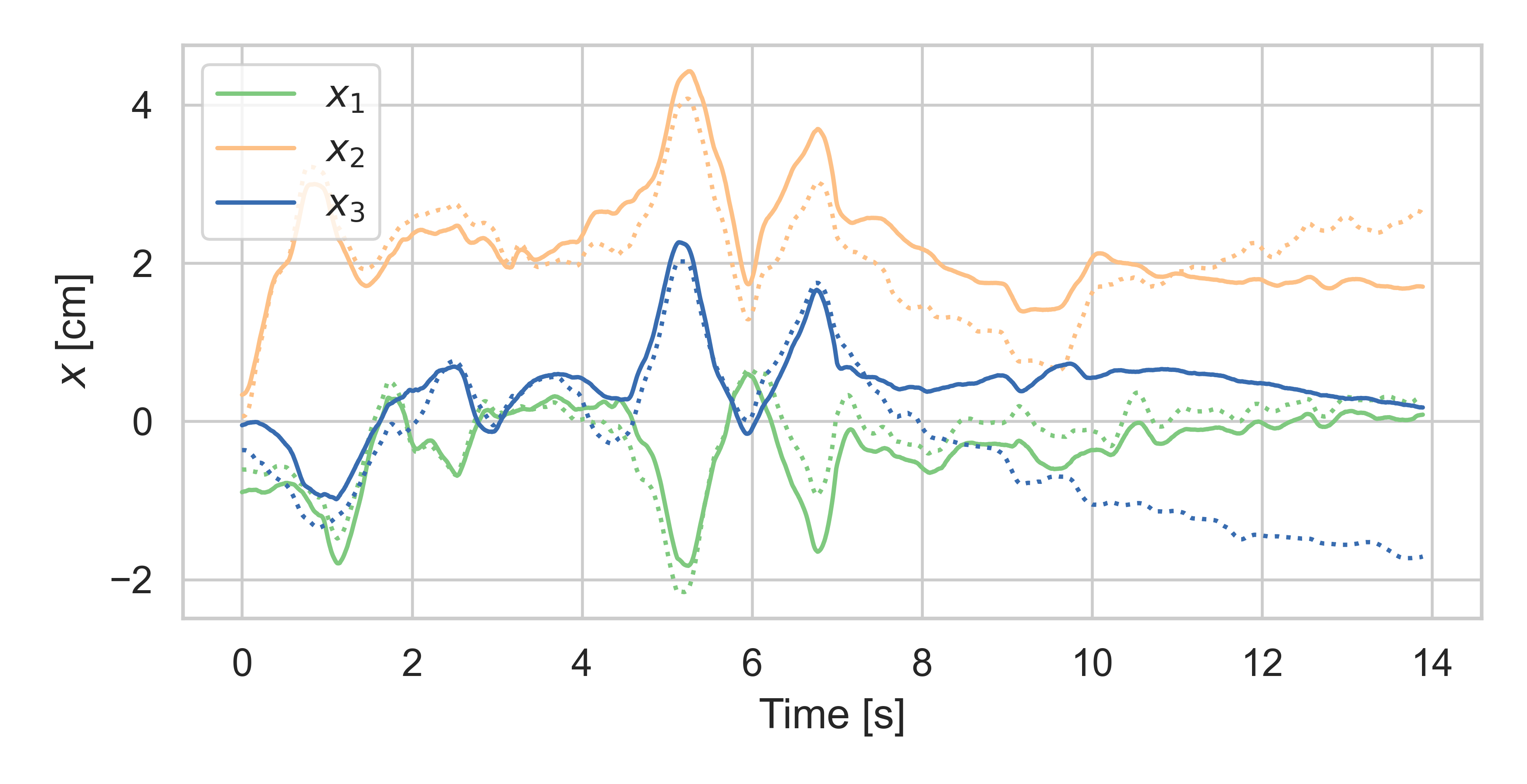}
    \vskip -0.45cm
    \caption{Prediction (solid lines) and ground truth (dotted lines) of cube position $x$ for an exemplary manipulation sequence.}
    \label{fig:filter_trajectory}
    \vskip -0.5cm
\end{figure}

During many manipulation episodes, the prediction error in the $x_3$-component accumulates over time as shown in \cref{fig:filter_trajectory}. This can be explained by the fact that the height of the position can not be uniquely determined by holding the cube on its four sides. Only when a finger reaches the upper edge of the cube (compare \cref{fig:setting_hand_cube}) conclusions on the $x_3$ position can be drawn.

\subsection{Real-World Experiments}

We employ the policy $\pi_1$, obtained from the iterative training procedure, on the real system.
We refer the reader to the accompanying video or the project website to see the robot successfully reaching each of the 24  $\pi/2$ goal orientations (\cref{fig:sequence} shows one example).
In~\cref{fig:24_goals}, we additionally plot the experimental success rate over 4 trials for each goal orientation.
Interestingly, the robot can reach many of the $24$ goal orientations very reliably, some of which showed lower success rates in simulation (across the whole domain randomization).
This means that for these movements the contact interactions are modeled well and lie in an area where the controller works.
On the other hand, some goal orientations, such as goal 6 are much more difficult for the real robot, while they are almost always solved in simulation.
This needs further analysis but we assume that the controller exploits certain interactions in the simulation that are not realistic/not modeled correctly.

\subsection{Spinning Friction}
\label{subsection:spinning_friction}

The spinning friction is particularly important to randomize for Sim2Real performance because, e.g., slight differences in spinning friction can yield qualitatively different behavior when the object is held by only two fingers (being stuck or swinging down).
Identifying the nominal values is difficult because they heavily depend on the concrete contacts.
Therefore, we randomize the friction in a wide range to ensure that the policy experiences these qualitative different behaviors and can find a robust strategy.

\begin{figure}
    \centering
    \includegraphics[width=0.9\linewidth]{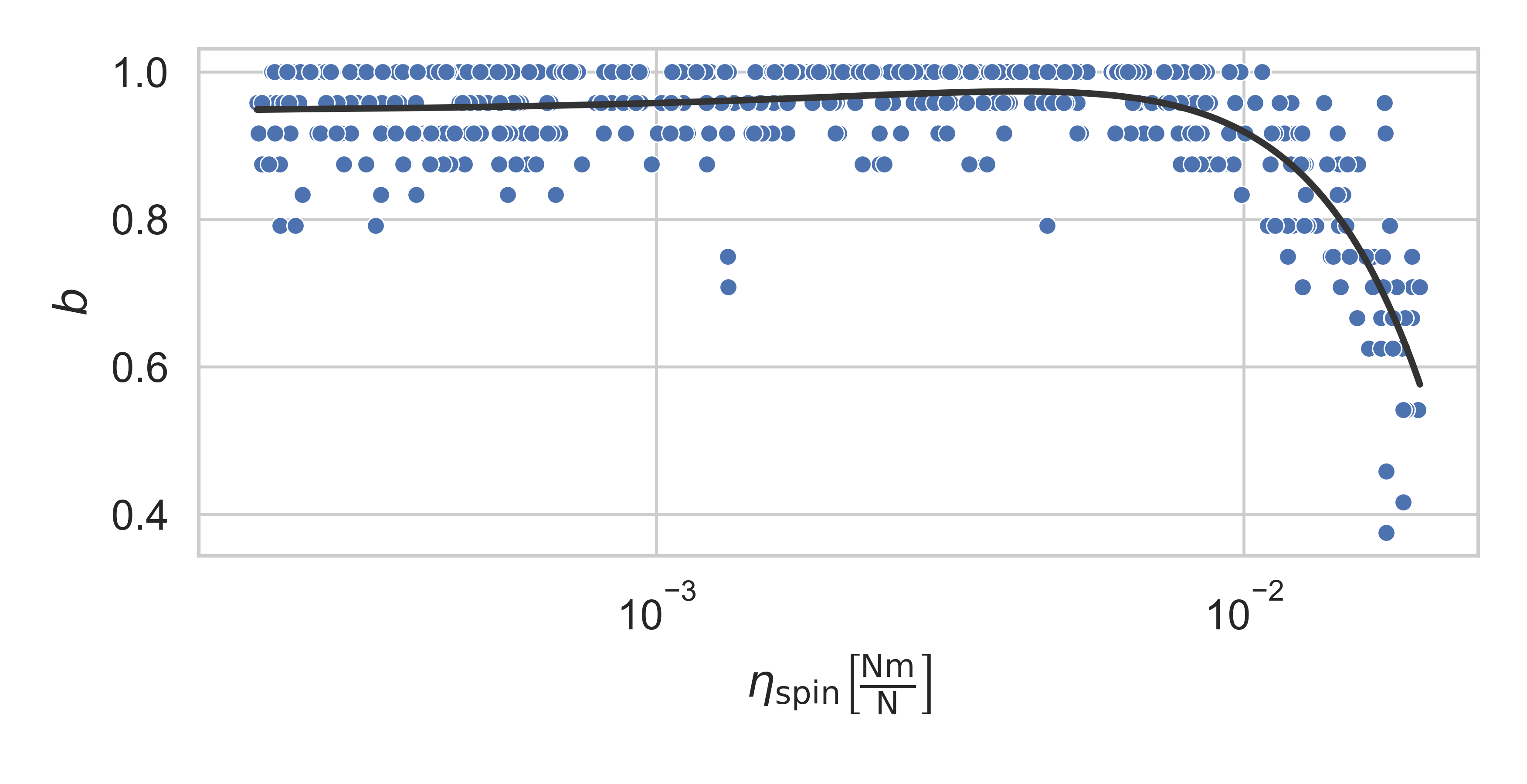}
    \vskip -0.4cm
    \caption{Mean success rate of the policy $\pi_1$ evaluated on the 24 goals benchmark with filter in-the-loop, plotted over a range of spinning friction values $\eta_\text{spin}$.}
    \vskip -0.5cm
    \label{fig:spinning_friction}
\end{figure}

In \cref{fig:spinning_friction}, we show that the wide range of values we expose the policy to makes the problem significantly harder in simulation.
Although we collected the data for the plot with the same sampling strategy as during the training, the policy performs significantly better at the lower end of the sampled friction values than at the upper end.
This shows that it is important to identify and simulate the system as precisely as possible to avoid that the policy focuses on unrealistic environment configurations if it cannot generalize across the whole domain.

\subsection{Discussion}

A key advantage of the modular learning approach we use is that the prediction of the estimator can be readily inspected and interpreted: during the development of the iterative training procedure, we were able to identify and address weaknesses of the estimator, the policy, and the interplay between the components.

During inference, we can inspect the predicted estimate (see \cref{fig:sequence}), which facilitates debugging the models in real-world settings.
Specifically, we found this to be crucial for identifying discrepancies between the simulation and the real system. We argue that the benchmark setting described in \cref{subsection:protocol} is particularly well suited for detecting modeling errors: the dexterous manipulation task requires precisely controlled movements of the fingers, and success depends on the intricacies of the contact-induced dynamics between the hand and the object (see \cref{subsection:spinning_friction}).

\section{Conclusions}

We have shown that purely tactile in-hand manipulation (with the hand upside down) for the task of reorienting a cube to a goal orientation is feasible.
In simulation, a success rate of 92\% (with significant domain randomization) is achieved, and via zero-shot Sim2Real-transfer to the torque-controlled DLR-Hand II, all 24 goal orientations could be reached with a high success rate.
For learning the task, we introduced a modular deep reinforcement learning architecture with two components: the policy for controlling the fingers which explicitly gets the cube state as an observation, and a deep differentiable particle filter for estimating this state from actual and desired joint angles (indirectly using the torque measurements).
This modularity allows for efficient stepwise training and detailed insight into the components (esp. via having an interpretable hidden state in the filter instead of a recurrent policy network).

In the future, we want to simplify the training procedure by utilizing adaptive domain randomization and make the policy even more robust by using the estimator uncertainty as additional input.
In addition, we want to integrate the learning of the policy and the filter in an end-to-end scheme to further increase performance.

\newpage
\footnotesize
\bibliographystyle{IEEEtranN-modified}

\bibliography{IEEEabrv, research}

\end{document}